\documentclass[final]{article}

\usepackage[dblblindworkshop]{neurips_2026}

\workshoptitle{The 6th Workshop on Mathematical Reasoning and AI}

\usepackage[utf8]{inputenc} 
\usepackage[T1]{fontenc}    
\usepackage{hyperref}       
\usepackage{url}            
\usepackage{booktabs}
\usepackage{array}       
\usepackage{amsfonts}       
\usepackage{nicefrac}       
\usepackage{microtype}      
\usepackage{xcolor}         

\usepackage{csquotes}
\usepackage{zref-clever}
\usepackage{amsmath,amssymb,amsthm}
\usepackage{graphicx}
\usepackage{tikz}
\usetikzlibrary{arrows.meta,positioning,fit,backgrounds,calc}

\usepackage{listings}

\lstdefinestyle{prompt}{
  basicstyle=\ttfamily\footnotesize,
  breakatwhitespace=true,
  breaklines=true,
  columns=fullflexible,
  frame=single,
  framexleftmargin=1.5em,
  keepspaces=true,
  numberblanklines=true,
  numbers=left,
  numbersep=8pt,
  numberstyle=\tiny\color{gray},
  showstringspaces=false,
  stepnumber=1,
  upquote=true,
  xleftmargin=2em
}

\lstdefinestyle{lean}{
  basicstyle=\ttfamily\scriptsize,
  breaklines=true,
  breakatwhitespace=true,
  columns=fullflexible,
  keepspaces=true,
  frame=single,
  showstringspaces=false,
  upquote=true,
  escapeinside={(*@}{@*)}
}

\usepackage{booktabs}
\usepackage{amsmath}
\usepackage{amssymb}
\DeclareUnicodeCharacter{2115}{\ensuremath{\mathbb{N}}}   
\DeclareUnicodeCharacter{2124}{\ensuremath{\mathbb{Z}}}   
\DeclareUnicodeCharacter{03B4}{\ensuremath{\delta}}       
\DeclareUnicodeCharacter{2208}{\ensuremath{\in}}          
\DeclareUnicodeCharacter{2211}{\ensuremath{\sum}}         
\DeclareUnicodeCharacter{2264}{\ensuremath{\leq}}         
\DeclareUnicodeCharacter{2265}{\ensuremath{\geq}}         
\DeclareUnicodeCharacter{2260}{\ensuremath{\neq}}         
\DeclareUnicodeCharacter{2261}{\ensuremath{\equiv}}       
\DeclareUnicodeCharacter{2286}{\ensuremath{\subseteq}}    
\DeclareUnicodeCharacter{2192}{\ensuremath{\rightarrow}}  
\DeclareUnicodeCharacter{00B7}{\ensuremath{\cdot}}        
\DeclareUnicodeCharacter{27E6}{\ensuremath{[\![}}         
\DeclareUnicodeCharacter{27E7}{\ensuremath{]\!]}}         
\DeclareUnicodeCharacter{27E8}{\ensuremath{\langle}}      
\DeclareUnicodeCharacter{27E9}{\ensuremath{\rangle}}      
\DeclareUnicodeCharacter{2014}{---}                       

\newcommand{\qwen}{Qwen3.8-27B}
\newcommand{\pos}[1]{\textbf{#1}}
\newcommand{\snd}[1]{\underline{#1}}   

\title{Improving semantic equivalence in research-level math formalization}

\newif\ifcameraready
\camerareadytrue   

\author{%
  Ayush Khaitan\\
  Princeton Language and Intelligence\\
  Princeton University\\
  Princeton, NJ 08540 \\
  \texttt{ayush.khaitan@princeton.edu} \\
  \And
  Liam Fowl\\
  Princeton Language and Intelligence\\
  Princeton University\\
  Princeton, NJ 08540 \\
  \texttt{lf2728@princeton.edu} \\
  \And
  Tomas Ortega\\
  Princeton Language and Intelligence\\
  Princeton University\\
  Princeton, NJ 08540 \\
  \texttt{tomas.ortega@princeton.edu} \\
  \And
  Alex Kontorovich\\
  Department of Mathematics\\
  Rutgers University\\
  Piscataway, NJ 08854\\
  \texttt{alex.kontorovich@rutgers.edu}
  \And
  Sanjeev Arora\\
  Princeton Language and Intelligence\\
  Princeton University\\
  Princeton, NJ 08540 \\
  \texttt{arora@princeton.edu} \\
}

\usepackage{xcolor}

\begin{document}

\maketitle

\begin{abstract}
    Formalizing mathematics in Lean is never a word-for-word translation of a paper: a Lean formalization changes definitions, strengthens hypotheses, splits arguments, and adds lemmas the paper never mentions. It stands to reason that a model that is good at autoformalization must be a good predictor of how a natural language source must differ from its faithful Lean formalization. With that in mind, we collect 4{,}076 audited examples of such changes from 832 Lean projects that formalize research-level mathematics, and use them to ask a practical question: how well can a model \emph{write} Lean inside a real research-level project that may contain tens of thousands of lines of code, and does training on discrepancies between natural language sources and their faithful Lean formalizations help? Our training set consists of 3{,}431 tasks from 141 repositories, and the test set consists of 645 tasks from the remaining 27 repositories. Training on in-project Lean nearly doubles complete proofs ($11\% \to 20\%$) and raises the share of statements in gold agreement with the project's own formalization by nearly a third ($30\%$ to $39\%$). The gain comes from the pairs themselves: an adapter trained only to predict how the Lean will differ, or a model told the difference in advance, barely helps. Training on where formalizations depart from their sources is thus an effective way of improving semantic equivalence in research-level formalization.

\end{abstract}

\section{Introduction}

When a mathematician formalizes a paper in Lean, the code may differ substantially from the natural language counterpart. A definition is generalized so that a Mathlib lemma applies; a hypothesis is added that was silently assumed by the paper; the Mathlib API doesn't support the proof given in the text, and hence the model must find a convoluted workaround which is sometimes an entirely different proof, etc. During autoformalization, predicting how the Lean formalization may differ from the natural language source is often the biggest time sink in any workflow; even frontier models like ChatGPT 5.6 Sol and Claude Fable often spend millions of tokens first trying to translate the natural language source verbatim into Lean, realize that this is impossible, and then either give up or spend millions more tokens trying to find elaborate workarounds. It stands to reason that models that are better at predicting how the Lean formalization will differ from its natural language counterpart will be much faster and much more efficient in formalization.

There is a second difficulty. Almost everything we know about language models and Lean comes from benchmarks of short, self-contained problems: competition exercises whose statement is handed to the model~\citep{zheng2022minif2f, tsoukalas2024putnambench}, or single sentences translated in isolation~\citep{wu2022autoformalization}. Formalizing a paper is a different activity. It produces a project of hundreds of declarations spread across many files, which builds up its own definitions and notation as it goes, so each new declaration has to fit the ones above. Research-level autoformalization therefore requires models to work within large projects, using substantial context from both the informal paper and the existing Lean code.

We study both difficulties in this paper. We first build a dataset of \emph{differences} between Lean formalizations and their informal sources: 4{,}076 records, each tied to a specific line in a Lean project and a specific sentence in the informal source. These records were isolated from 832 Lean repositories publicly available online, of which we retained only 171 after eliminating the rest for reasons such as incomplete code and misformalization. We then test whether training on examples from this dataset improves autoformalization. Throughout, we focus on writing Lean inside large repositories, where a declaration must fit the surrounding project.

In particular, we fine-tune the base model \qwen{} with a LoRA adapter, and compare it against the untouched model, as well as several other models in the same class. The adapter is trained on the Lean declarations associated with the recorded differences. It is shown the project file up to the point of interest and the informal passage, and must output the declaration the project actually wrote. The adapter does not see the Lean compiler during training.

\paragraph{Research questions.}
\begin{itemize}
\item How well do current open models write Lean inside a large research project, when the task comes from research-level mathematics?
\item Does LoRA fine-tuning \qwen{} on examples of differences between informal mathematics and Lean from other projects improve its writing?
\end{itemize}

\section{The benchmark}

\paragraph{Where the tasks come from.}
Every entry in the dataset points at a declaration, i.e. a \texttt{def}, \texttt{theorem}, \texttt{lemma}, or \texttt{structure} at a pinned commit, and at the passage in the natural language paper that it formalizes. We cloned the Lean projects at those commits and then built them locally. We then held out 28 projects that were never used for any fine-tuning in this paper, and kept only declarations whose original code still compiled in our own checking harness. This left 645 tasks from 27 projects, with at most 30 from any single project.

\paragraph{What a task looks like.}
For each declaration we cut the file at that point. The model sees everything above the cut (imports and the project's earlier definitions) plus the informal passage, and is asked for one of three things: \textbf{def} (145 tasks), write the definition; \textbf{thm} (250 tasks), write the theorem statement, ending in \texttt{:= by sorry}; \textbf{proof} (250 tasks), given the exact statement, write the proof. 

\paragraph{How answers are scored.}
Every answer is pasted into the original file at the cut and compiled with the project's own Lean and Mathlib versions. All models are given the same tasks. We report three quantities:
\begin{itemize}
    \item \textbf{Compiles}: the code elaborates without errors (for proofs, also without \texttt{sorry}). For proofs this is the whole score, since the statement is given and a proof that compiles is correct.
    \item \textbf{Natural language (NL) agreement}: for definitions and statements, compiling is not enough, since \texttt{theorem foo : True := by sorry} compiles. A judge model reads the paper passage, the project's own declaration, and the candidate Lean declaration, and scores whether the candidate's Lean states the mathematical content of the passage correctly, allowing reasonable formal choices. NL agreement means it compiles \emph{and} gets the top score. This is what autoformalization benchmarks usually measure. Note that this is an imperfect measure. For example, the natural language statement may omit a hypothesis that is clear from the context of the paper, but must still be mentioned in the formalized statement. The formalization receives a full score from ChatGPT 5.6 Sol, although it may have skipped putting in this hypothesis. That is why we have a stricter measure that we call ``Gold agreement"; it is defined below.
    \item \textbf{Gold agreement}: the judge also scores whether the candidate is the same mathematical statement as the one the project actually wrote, up to naming, hypothesis order, and definitional unfolding. Gold agreement means it compiles and gets the top score here. This is a different question from NL agreement, for two reasons. The project itself departs from the paper---that is the premise of the dataset---so a statement that transcribes the passage can still be the wrong theorem for the project, for instance because it lacks a hypothesis the project had to add. And a passage usually covers several declarations, so a candidate identical to the project's lemma can be scored as stating less than the passage. Gold agreement is the stricter measure and the one the task poses, so we treat it as primary for statements.
\end{itemize}

Note that gold agreement is a stricter and more reliable test of semantic equivalence than NL agreement. Examples of incorrect formalizations that are incorrectly judged to be in NL agreement, but correctly judged to not be in gold agreement, are given in the appendix. We will primarily focus on gold agreement in this paper.

The judge is \texttt{gpt-5.6-sol} at temperature $0$; gpt-oss-120b serves as a second judge. They receive the informal source, the project's gold declaration, the project's recorded note on how its formalization deliberately differs from the source, the candidate, and whether the candidate compiled, and returns both scores, each in $\{0,1,2\}$, in a single JSON object (in the released code the fields are named \texttt{faithful} and \texttt{equivalent\_to\_gold}; they are our NL agreement and Gold agreement respectively). They are instructed to judge mathematical content rather than style, and explicitly not to penalize differences in names, hypothesis order, or notation. Both measures binarise at the top score, so a candidate judged ``close but with a substantive difference in hypotheses, scope or representation'' counts as a failure. Uncertainty is a paired bootstrap over tasks (95\% percentile intervals, 10{,}000 resamples), so conditions are always compared on the same tasks.

Note that the two judges differ substantially on NL agreement, but almost always agree on gold agreement; this is another reason why gold agreement is a better measure to study in this paper. In this paper we will use ``semantically equivalent" and ``in gold agreement" interchangeably.

\section{Models}

All seven models are run with greedy decoding, one sample per task, through vLLM on a single 80\,GB GPU. \textbf{Untuned:} \qwen{} (our base, thinking disabled), Qwen3-32B, Gemma-3-27B, gpt-oss-120b, DeepSeek-R1-Distill-Qwen-32B, and Goedel-Prover-32B~\citep{lin_goedel-prover-v2_2025}, trained specifically for Lean proving. \textbf{\qwen{} + writer adapter:} a rank-16 LoRA trained on the writing task itself---3{,}256 examples of (file prefix, informal passage) $\to$ (Lean declaration) from dataset projects absent from the benchmark, 2 epochs, $\approx$8.5 GPU-hours. It is supervised fine-tuning with loss on completion tokens only; it does not see compiler feedback during training.

\section{Results}

\begin{table}[t]
\centering
\caption{Results over the full task count ($n = 145/250/250$ for def/thm/proof; gpt-oss-120b has $119/241/250$ because 26 definition and 9 statement generations returned no code; Appendix~\ref{app:gap} gives its rates over the full counts). NL agreement: compiles and is judged to state the paper's passage correctly. Gold agreement: compiles and is judged the same statement as the project's own declaration. For proofs the statement is given, so Compiles is the score. Best per column in bold, second best underlined (ties share the mark).}
\label{tab:main}
\small
\begin{tabular}{l ccc cc cc}
\toprule
 & \multicolumn{3}{c}{Compiles} & \multicolumn{2}{c}{NL agreement} & \multicolumn{2}{c}{Gold agreement} \\
\cmidrule(lr){2-4}\cmidrule(lr){5-6}\cmidrule(lr){7-8}
Model & def & thm & proof & def & thm & def & thm \\
\midrule
\qwen{}                & .476 & \snd{.744} & .112 & \snd{.166} & \pos{.360} & .131 & .304 \\
\quad + writer adapter & .476 & .700 & .204 & \snd{.166} & \snd{.356} & \snd{.166} & \pos{.392} \\
Qwen3-32B              & .255 & .580 & .096 & .103 & .256 & .110 & .220 \\
Gemma-3-27B            & .276 & .504 & .068 & .076 & .168 & .062 & .164 \\
gpt-oss-120b           & \snd{.529} & .647 & \snd{.224} & \pos{.227} & .332 & \pos{.210} & \snd{.332} \\
R1-Distill-Qwen-32B    & .317 & .564 & .112 & .103 & .220 & .083 & .196 \\
Goedel-Prover-32B      & \pos{.566} & \pos{.784} & \pos{.244} & .124 & .304 & .083 & .288 \\
\bottomrule
\end{tabular}
\end{table}

\paragraph{LoRA on difference pairs improves gold agreement and proofs.}
The writer adapter wins on two independent measures. It nearly doubles proof completion, $11.2\% \to 20.4\%$ ($+9.2$ points, 95\% interval $[+4.8, +13.6]$), and it raises the share of statements in gold agreement from $30.4\%$ to $39.2\%$ ($+8.8$ points, $[+3.6, +14.0]$), the highest of any model we tested and $6.6$ points ($[+1.2, +12.0]$) above gpt-oss-120b. Its statements compile slightly less often ($-4.4$, $[-10.0, +0.8]$), but those that compile are in gold agreement far more often (56\% against 41\% for the base). On definitions it changes nothing significant (compiles $0.0$, $[-8.3, +7.6]$; NL agreement unchanged; gold agreement $+3.4$, $[-1.4, +9.0]$), and on statements it leaves the NL agreement rate where it was ($-0.4$, $[-5.6, +4.8]$). Also note that the fine-tuned model is either the best or second-best across the board under both the ``NL agreement" and ``gold agreement" measures. A detailed table with the corresponding 95\% confidence intervals is in~\ref{tab:paired}.

This is perhaps the standout takeaway of this experiment: although current open source models are fairly good at writing Lean statements that compile without errors, their ability to write semantically correct research-level math statements leaves much to be desired. In fact the gap is so large that even though LoRA fine-tuning made the model slightly worse at writing statements that compile, it is still able to raise the share of gold agreement statements by nearly a third ($30\% \to 39\%$); and on proofs, where compiling is the whole task, it nearly doubled completion ($11.2\% \to 20.4\%$). Appendix~\ref{app:gap} gives all paired differences.

\paragraph{Why NL agreement and Gold agreement disagree.}
The adapter raises the Gold agreement score on statements by $8.8$ points but leaves the NL agreement score where it was, so it is worth looking at how the two scores overlap. Among compiling statements, the base model has 59 candidates that are both in NL agreement and in gold agreement, 31 that are in NL agreement but not in gold agreement, and 17 that are in gold agreement but not in NL agreement; for the adapter the counts are 72, 17 and 26. The number of statements in NL agreement barely changes (90 against 89). What changes is the middle group. These are candidates that restate the paper correctly but are not the theorem the project wrote: the judge's notes cite a hypothesis the project added, a bound stated pointwise where the project states it eventually, or a single claim where the project proves several at once. They miss exactly the kind of departure our dataset records, and a judge that only compares against the paper cannot reward a candidate for making it. The adapter turns fourteen of these into exact matches of the project's statement. The third group, in gold agreement but not in NL agreement, looks odd at first, since a statement identical to the project's is a correct formalization by definition. The judge's notes explain it: the passage often describes more than the one declaration (a helper lemma is judged against a whole theorem, or one lemma against a paragraph that proves several), so NL agreement penalizes a mismatch in granularity that has nothing to do with the candidate. This happens more often for the adapter precisely because more of its statements match the project exactly. In short, fine-tuning on the difference pairs does not make the model better at restating the paper; it makes the model make the same formalization choices the project made.

\paragraph{LoRA fine-tuning on difference labels does not improve performance}
We also tried the following: we made a list of 17 descriptive labels to denote the different ways in which the Lean formalization differs from its natural language counterpart, and then trained a LoRA \textit{gap adapter} on Qwen3.8-27B on just those labels. We wanted to check if being trained on those labels would help with autoformalization. The gap adapter is shown the informal passage, never any Lean, and must output the label together with a one-line description of the change and why the formalizer would make it (920 records, 3 epochs, $\approx$2 GPU-hours). When asked to output Lean code, it helps on neither measure: its proof gain is not significant ($+3.2$ points, $[-0.4, +6.8]$), its statement gold-agreement rate is flat ($+0.4$), and it compiles significantly less code for both definitions ($-8.3$) and statements ($-7.2$). Second, we re-ran every model with the labels supplied in the prompt, either the true labels or those the gap adapter predicts; this changes scores by a few points in no consistent direction, and the writer adapter stays ahead of the base model under both. Knowing \emph{about} the gap, in this form, does not transfer to closing it. Full numbers are in Appendix~\ref{app:gap}.

\paragraph{Writing in a real project is difficult for all models.}
Table~\ref{tab:main} shows the results. The best model completes about one proof in four, and on statements no model exceeds 36\% NL agreement or 39\% gold agreement. Compiling and being right are far apart: Goedel-Prover compiles the most statements (78\%) but only 30\% are in NL agreement and 29\% in gold agreement. Model size and general capability matter less than one might expect. A 27B model with a small amount of in-project training is competitive with gpt-oss-120b, which has 120B parameters.



\section{Conclusion}

We built a dataset of 4{,}076 (natural language, Lean) examples from 171 research-level Lean projects, centred on the places where a formalization departs from its paper, and used it both to fine-tune \qwen{} with a LoRA adapter and to benchmark seven models on writing definitions, statements and proofs inside 27 held-out projects. The adapter nearly doubled proof completion and raised the share of statements in gold agreement with the project's own formalization by nearly a third, the best of the seven models, while compiling no more statements or definitions than the base model. The gain is therefore not a general improvement in Lean skill; the model has learned to make the formalization choices that projects actually make, which a literal transcription of the paper misses. The benchmark, adapters, both judge runs, and evaluation code are released with this paper.\footnote{\url{https://anonymous.4open.science/r/lean-writing-bench-CCB2/README.md}}

\begin{ack}
    The authors acknowledge funding from the DARPA expMath program. A.K.~acknowledges funding from the AMS Simons Travel grant. 
\end{ack}

\bibliographystyle{plainnat}
\bibliography{references}
\appendix

\section{Supplementary details}
\label{app:details}

\paragraph{How the dataset was built.}
We started from 832 public Lean repositories that formalize research-level mathematics. An audit removed repositories with, for example, incomplete code or misformalized results; 171 survived, and 168 of those contain at least one declaration that compiles in our checking harness. Within each surviving repository the audit recorded the places where the Lean departs from its informal source. One \emph{record} consists of the informal passage, the Lean declarations that formalize it, one or more labels from the 17 types in Table~\ref{tab:taxonomy}, and a short note saying what changed and why. There are 1{,}160 records, carrying 1{,}482 labels in total. A record can cover several declarations, so expanding records to declarations gives 2{,}721 (passage, declaration) pairs. The pairs from the 141 training repositories become the 3{,}431 examples used to train the writer adapter (1{,}348 statements, 1{,}348 proofs and 735 definitions; 3{,}256 for training and 175 for validation). The pairs from the 27 held-out repositories become the 395 benchmark declarations and their 645 tasks. Adding the two gives the 4{,}076 examples quoted in the abstract. The split is by repository, so no training example comes from a benchmark project. Every writer training example therefore sits at a place where the project's Lean is known to differ from the paper; the model is trained to write the Lean the project wrote there, not the label describing the difference.

\begin{table}[h]
\centering
\caption{The 17 difference types and how many of the 1{,}160 records carry each label (a record can carry more than one).}
\label{tab:taxonomy}
\small
\begin{tabular}{lr@{\quad}lr}
\toprule
Type & $n$ & Type & $n$ \\
\midrule
proof\_detail\_expansion & 319 & scope\_change & 58 \\
statement\_change & 185 & source\_correction & 53 \\
proof\_strategy\_change & 175 & auxiliary\_result\_or\_construction & 42 \\
representation\_change & 109 & dependency\_or\_type\_system\_accommodation & 33 \\
hypothesis\_change & 97 & constructivity\_or\_computational\_change & 29 \\
omission\_or\_placeholder & 91 & formal\_scaffolding & 27 \\
definition\_or\_interface\_change & 83 & totalization\_or\_choice & 21 \\
case\_analysis\_expansion & 76 & misformalization & 11 \\
construction\_change & 73 & & \\
\bottomrule
\end{tabular}
\end{table}

\paragraph{Held-out projects.} The 27 benchmark repositories are dataset entries 0145, 0179, 0262, 0292, 0421, 0423, 0424, 0463, 0465, 0469, 0470, 0483, 0487, 0509, 0511, 0563, 0564, 0612, 0618, 0621, 0627, 0637, 0642, 0669, 0670, 0674, and 0767. A 28th held-out project had no declaration that passed the gold check and contributes no tasks. The projects contain between about 2{,}100 and 113{,}000 lines of Lean source each (median about 10{,}600; 14 of the 27 have more than 10{,}000 lines). Declaration kinds: 171 theorems, 116 definitions, 79 lemmas, 13 structures, 8 classes, 7 abbreviations, 1 inductive.

\paragraph{Prompt.} The system message is: ``You are an expert Lean 4 / Mathlib formalizer contributing to an existing research formalization project. You are given the beginning of a Lean file from that project (its imports and preceding declarations) and a natural-language mathematical source. Write Lean 4 code that fits that file: use the project's own definitions and notation from the prefix where relevant, and current Mathlib names. Output only a single \texttt{```lean} code block containing exactly what is asked for, with no commentary.'' The user message contains the file prefix (imports preserved, middle elided if over 6{,}000 characters), the informal source (identifier, paraphrase, excerpt where available, and the informal argument), the difference-label hint in the hinted settings of Appendix~\ref{app:gap}, the statement to prove for the proof task, and a per-task instruction. The writer adapter was trained on this same format with \texttt{context = none}, i.e. no hint.

\paragraph{Fine-tuning.} All adapters are rank-16 LoRA ($\alpha = 32$, dropout 0.05) on all text-side linear projections of \qwen{} (0.4\% of parameters), learning rate $10^{-4}$, effective batch 8, bf16, cross-entropy on completion tokens only with prompt tokens masked to $-100$. Writer adapter: 3{,}256 training / 175 validation examples split by repository, 2 epochs, max length 4{,}096, final validation loss 0.55. Gap adapter: 920 training / 240 validation records (repository-split), 3 epochs, max length 3{,}072. Size-matched writer adapter (Appendix~\ref{app:gap}): a random 920 of the 3{,}256 writer training examples (179 definitions, 357 statements, 384 proofs), 3 epochs, otherwise identical to the writer adapter.

\paragraph{Serving.} vLLM with greedy decoding and a 32k-token context; 32B models use an fp8 KV cache to fit one 80\,GB GPU. Proof tasks allow up to 12k output tokens for reasoning models.

\section{Limitations}
\label{app:limits}

The benchmark has 395 declarations from 27 projects, so confidence intervals are wide, especially for definitions. Two projects are standalone files that we compiled inside a host project with a slightly different Mathlib; many of their items failed our gold check and were dropped. NL agreement and Gold agreement are judged by a language model, not by humans. We ran a second judge (Appendix~\ref{app:judge}) and read a sample of verdicts by hand, but we did not audit all 9{,}022. Every model was run greedily, with one sample and no compiler feedback, so the absolute proof rates are below what an agentic or search-based system would reach; our interest is in the differences between conditions, which all share this setup. The held-out projects are public, so the untuned models may have seen them during pretraining; what we can guarantee is that no model was \emph{fine-tuned} on them. We checked for memorization directly: no model reproduces any of the 250 gold statements word for word, and only 2--6 of the 145 gold definitions are reproduced word for word by any one model, the writer adapter included. The writer adapter was trained on declarations that sit at recorded departures from the paper; we did not train a control on declarations from the same projects that are \emph{not} at such places, so we cannot say how much of the gain comes from choosing those examples and how much from in-project Lean data in general. Appendix~\ref{app:gap} does show that the gain is not explained by the amount of training data alone. Only one base model was fine-tuned, and the adapter was small and trained by supervised fine-tuning alone; the obvious next step, reinforcement learning against the compiler, is exactly what we did not do.

\section{Robustness to the judge}
\label{app:judge}

Two of our three metrics depend on a model's opinion, so we ran the entire evaluation under a second judge, the locally hosted gpt-oss-120b. On the 9{,}022 judged candidates the two judges agree on $94.0\%$ of Gold agreement decisions ($\kappa = 0.84$) and $83.2\%$ of NL agreement decisions ($\kappa = 0.57$). The Gold agreement marginals are almost identical ($25.8\%$ versus $25.8\%$), so every Gold agreement number and the writer adapter's $+8.8$ are insensitive to the choice. NL agreement is where the judges differ: gpt-oss-120b is uniformly stricter, scoring $22.7\%$ of candidates in NL agreement against $30.0\%$ for gpt-5.6-sol, so all NL agreement numbers shift down by roughly seven points under it. The ranking of models is unchanged. This also lets us check a conflict of interest in the earlier setup, where gpt-oss-120b was simultaneously the judge and one of the evaluated models: its own NL agreement rate rises by $8.4$ points under the independent judge, but so does every other model's, by $3.5$ to $9.2$ points, and the largest gain goes to the writer adapter rather than to gpt-oss-120b. We find no evidence that the model favoured its own outputs. Both judge runs are released.

\paragraph{What the judge is asked.} The system message is: ``You are a careful referee for Lean 4 formalizations of research mathematics. You compare a candidate Lean declaration against (a) the informal source it should formalize and (b) the declaration the project actually wrote (gold), together with the project's recorded note on how its formalization deliberately differs from the informal source. Judge mathematical content, not style or naming. Respond with a single JSON object and nothing else.'' The user message gives the informal source, the gold declaration (with the instruction ``do not penalize the candidate for differing in names, order of hypotheses, or notation''), the project's recorded design decision, the candidate and whether it compiled, and asks for \texttt{\{"faithful": 0|1|2, "equivalent\_to\_gold": 0|1|2, "gap\_realized": 0|1, "notes": ...\}}. A score of 2 on \texttt{faithful} means ``states the informal content correctly (allowing reasonable formal choices)''; a score of 2 on \texttt{equivalent\_to\_gold} means ``same mathematical statement/object as the gold (up to naming, hypothesis order, definitional unfolding)''; a score of 1 on either means ``close but with a substantive difference in hypotheses, scope or representation''. NL agreement is \texttt{faithful} $= 2$ and Gold agreement is \texttt{equivalent\_to\_gold} $= 2$, both restricted to candidates that compile.

\paragraph{Two verdicts, one from each side.} The two examples below are real base-model candidates and the judge's own notes on them. They show why the two scores can disagree.

\emph{In NL agreement but not in gold agreement.} The passage says only that the paper ``uses elementary log-factorial bounds''. The project's lemma is
\begin{quote}\small
\texttt{factorial\_le\_pow (k : ℕ) (hk : 1 ≤ k) :}\\
\texttt{\quad ((2 * k).factorial : $\mathbb{R}$) ≤ (2 * k : $\mathbb{R}$) \^{} (2 * k)}
\end{quote}
The candidate is
\begin{quote}\small
\texttt{factorial\_le\_pow (n : ℕ) : (n! : $\mathbb{R}$) ≤ (n : $\mathbb{R}$) \^{} n}
\end{quote}
Judge's note: ``the same factorial-to-power estimate in a valid stronger form for all naturals, whereas the gold restricts to positive $k$ and the even input $2k$.'' NL agreement 2, Gold agreement 1. The candidate is a correct formalization of the sentence, but it is not the lemma the project wrote, which is stated only for $(2k)!$ with $k \geq 1$.

\emph{In gold agreement but not in NL agreement.} The passage describes a scaling identity, $F(bx) = \cos^2(2\pi x)\,F(x)$. At that point the project's declaration is only a small helper lemma,
\begin{quote}\small
\texttt{F\_nonneg (b : ℕ) ($\xi$ : $\mathbb{R}$) : 0 ≤ F b $\xi$}
\end{quote}
and the candidate is the same statement with \texttt{b} made implicit. Judge's note: ``mathematically identical to the gold up to making $b$ implicit, but it states only nonnegativity of $F$ and neither expresses nor internally proves the cited scaling identity.'' NL agreement 1, Gold agreement 2. The candidate matches the project exactly, but the passage covers more than this one declaration, so a judge that compares only against the passage marks it down.

\section{The gap adapter, difference-label hints, and controls}
\label{app:gap}

\paragraph{The gap adapter.}
The gap adapter is a LoRA trained on the difference records themselves, with the held-out projects removed. It never sees Lean. Its input is a paraphrase of the informal passage and a one-line formalization goal, and its output is one of the 17 labels together with a few sentences describing the change (Appendix~\ref{app:examples} shows a training example). On the held-out projects it scores 0.29 micro-F1, which is also what always guessing the most common label scores; on unseen projects it has learned essentially the label frequencies. A split that does not hold out whole projects gives much higher scores, which is why we report the repository-level split.

\paragraph{Is the writer adapter's advantage just more training data?}
The gap adapter has fewer examples (920) and a different recipe (3 epochs) than the writer adapter (3{,}256 examples, 2 epochs), so its failure could simply be a matter of size. To check, we trained a third adapter with the gap adapter's recipe but the writer adapter's kind of data: a random 920 of the 3{,}256 writer examples, 3 epochs, all other settings unchanged. Table~\ref{tab:ctrl} compares the four models. With the same amount of data and the same recipe, learning to write the Lean beats learning to name the difference: the 920-example writer adapter gains $+6.0$ proof points over the base model ($[+2.0, +10.0]$) where the gap adapter gains $+3.2$ ($[-0.4, +6.8]$), and $+2.4$ statement gold-agreement points ($[-2.4, +7.2]$) where the gap adapter gains $+0.4$. Going from 920 to 3{,}256 examples then raises the statement gold-agreement gain from $+2.4$ to $+8.8$ and the proof gain from $+6.0$ to $+9.2$. So the writer adapter's advantage is not an artefact of its size, and it grows with more data rather than saturating.

\begin{table}[h]
\centering
\caption{Size-matched control, no hint, in the format of Table~\ref{tab:main}. ``writer adapter (920)'' uses the gap adapter's data size and recipe.}
\label{tab:ctrl}
\small
\begin{tabular}{l ccc cc cc}
\toprule
 & \multicolumn{3}{c}{Compiles} & \multicolumn{2}{c}{NL agreement} & \multicolumn{2}{c}{Gold agreement} \\
\cmidrule(lr){2-4}\cmidrule(lr){5-6}\cmidrule(lr){7-8}
Model & def & thm & proof & def & thm & def & thm \\
\midrule
\qwen{}                      & .476 & .744 & .112 & .166 & .360 & .131 & .304 \\
\quad + gap adapter          & .393 & .672 & .144 & .172 & .324 & .124 & .308 \\
\quad + writer adapter (920) & .503 & .696 & .172 & .145 & .324 & .159 & .328 \\
\quad + writer adapter       & .476 & .700 & .204 & .166 & .356 & .166 & .392 \\
\bottomrule
\end{tabular}
\end{table}

\paragraph{Three amounts of hint.}
Each task is posed three ways: with \emph{no} hint (the setting of Table~\ref{tab:main}); with a \emph{predicted} hint, the difference types the gap adapter outputs for the passage; and with the \emph{oracle} hint, the true difference types from the dataset. A prediction exists for 148 of the 164 passages, so the predicted setting has slightly fewer tasks. Table~\ref{tab:paired} gives the writer adapter's paired gains in all three settings, and Table~\ref{tab:hints} gives every model's scores under both hints.

\begin{table}[h]
\centering
\caption{Paired differences of the writer adapter against base \qwen{} (percentage points) with 95\% bootstrap intervals, in the three hint settings. Positive is better; bold when the interval excludes zero.}
\label{tab:paired}
\small
\begin{tabular}{l r@{\ }l r@{\ }l r@{\ }l}
\toprule
Metric & \multicolumn{2}{c}{no hint} & \multicolumn{2}{c}{predicted hint} & \multicolumn{2}{c}{oracle hint} \\
\midrule
proof compiles & \pos{$+9.2$} & $[+4.8, +13.6]$ & $+4.1$ & $[-0.9, +8.6]$ & \pos{$+6.0$} & $[+2.0, +10.0]$ \\
thm Gold agreement  & \pos{$+8.8$} & $[+3.6, +14.0]$ & \pos{$+6.8$} & $[+1.4, +11.8]$ & $+4.8$ & $[-0.4, +10.0]$ \\
thm NL agreement   & $-0.4$ & $[-6.0, +4.8]$ & $+1.8$ & $[-3.6, +6.8]$ & $+1.2$ & $[-3.6, +6.4]$ \\
thm compiles  & $-4.4$ & $[-10.0, +0.8]$ & \pos{$-7.7$} & $[-13.6, -2.3]$ & \pos{$-7.6$} & $[-13.2, -2.0]$ \\
def compiles   & $0.0$ & $[-8.3, +7.6]$ & $+3.1$ & $[-5.4, +12.3]$ & $+2.8$ & $[-5.5, +11.0]$ \\
\bottomrule
\end{tabular}
\end{table}

\begin{table}[h]
\centering
\caption{All seven models under the two hint settings, in the format of Table~\ref{tab:main}. Rates are over the tasks posed in each setting: $145/250/250$ (def/thm/proof) with the oracle hint and $130/220/220$ with the predicted hint; gpt-oss-120b has $125/240/250$ and $109/208/220$ because some of its generations returned no code. For reference, scoring its missing generations as failures gives gpt-oss-120b no-hint rates of .434 compiles, .186 NL agreement and .172 Gold agreement on definitions, and .320 Gold agreement on statements, over the full $145/250$.}
\label{tab:hints}
\small
\begin{tabular}{l ccc cc cc}
\toprule
 & \multicolumn{3}{c}{Compiles} & \multicolumn{2}{c}{NL agreement} & \multicolumn{2}{c}{Gold agreement} \\
\cmidrule(lr){2-4}\cmidrule(lr){5-6}\cmidrule(lr){7-8}
Model & def & thm & proof & def & thm & def & thm \\
\midrule
\multicolumn{8}{l}{\emph{predicted hint}} \\
\qwen{}                      & .462 & .745 & .127 & .138 & .318 & .123 & .295 \\
\quad + writer adapter       & .492 & .668 & .168 & .123 & .336 & .146 & .364 \\
Qwen3-32B                    & .285 & .541 & .095 & .085 & .245 & .085 & .236 \\
Gemma-3-27B                  & .223 & .473 & .059 & .062 & .136 & .054 & .145 \\
gpt-oss-120b                 & .486 & .673 & .205 & .156 & .317 & .156 & .327 \\
R1-Distill-Qwen-32B          & .308 & .523 & .145 & .108 & .177 & .069 & .155 \\
Goedel-Prover-32B            & .508 & .764 & .205 & .138 & .295 & .092 & .273 \\
\midrule
\multicolumn{8}{l}{\emph{oracle hint}} \\
\qwen{}                      & .441 & .768 & .124 & .166 & .376 & .145 & .356 \\
\quad + writer adapter       & .469 & .692 & .184 & .172 & .388 & .200 & .404 \\
Qwen3-32B                    & .297 & .592 & .120 & .110 & .300 & .117 & .300 \\
Gemma-3-27B                  & .241 & .516 & .064 & .069 & .180 & .048 & .200 \\
gpt-oss-120b                 & .552 & .708 & .196 & .264 & .362 & .240 & .362 \\
R1-Distill-Qwen-32B          & .338 & .556 & .136 & .110 & .228 & .103 & .196 \\
Goedel-Prover-32B            & .476 & .792 & .208 & .117 & .328 & .103 & .304 \\
\bottomrule
\end{tabular}
\end{table}

\paragraph{Hints about the gap barely move the needle.}
Adding predicted or true difference types changes scores by a few points in no consistent direction, and no model shows a significant change in its definition or statement \emph{NL agreement} score from either hint. The significant effects are small and scattered: Qwen3-32B gains 2.8 proof points from the oracle hint, R1-Distill gains 4.1 from the predicted hint, and Goedel-Prover gets \emph{worse} (definitions $-9.0$ with oracle, proofs $-3.6$ with predicted), presumably because extra prose distracts a model tuned on bare Lean. If the true labels help this little, a predictor of those labels cannot help much. Table~\ref{tab:paired} shows that the writer adapter's gains survive the hints: both keep their sign under both hints, the Gold agreement gain stays significant under the predicted hint, and proof completion stays significant under the oracle hint.

\begin{table}[h]
\centering
\caption{The gap adapter used as a writer. Top: no-hint results in the format of Table~\ref{tab:main}. Bottom: paired differences against base \qwen{} (percentage points) with 95\% bootstrap intervals, bold when the interval excludes zero.}
\label{tab:gap}
\small
\begin{tabular}{l ccc cc cc}
\toprule
 & \multicolumn{3}{c}{Compiles} & \multicolumn{2}{c}{NL agreement} & \multicolumn{2}{c}{Gold agreement} \\
\cmidrule(lr){2-4}\cmidrule(lr){5-6}\cmidrule(lr){7-8}
Model & def & thm & proof & def & thm & def & thm \\
\midrule
\qwen{}                & .476 & .744 & .112 & .166 & .360 & .131 & .304 \\
\quad + gap adapter    & .393 & .672 & .144 & .172 & .324 & .124 & .308 \\
\bottomrule
\end{tabular}

\medskip

\begin{tabular}{l r@{\ }l r@{\ }l r@{\ }l}
\toprule
Metric & \multicolumn{2}{c}{no hint} & \multicolumn{2}{c}{predicted hint} & \multicolumn{2}{c}{oracle hint} \\
\midrule
proof compiles & $+3.2$ & $[-0.4, +6.8]$ & $+0.5$ & $[-4.1, +5.0]$ & $+1.6$ & $[-1.6, +4.8]$ \\
thm Gold agreement  & $+0.4$ & $[-3.6, +4.4]$ & $-1.8$ & $[-5.9, +2.3]$ & $-3.6$ & $[-8.0, +0.8]$ \\
thm NL agreement   & $-3.6$ & $[-8.0, +0.8]$ & $-4.1$ & $[-8.6, +0.5]$ & $-3.6$ & $[-8.0, +0.8]$ \\
thm compiles  & \pos{$-7.2$} & $[-12.0, -2.4]$ & \pos{$-10.9$} & $[-16.4, -5.9]$ & \pos{$-10.0$} & $[-14.8, -4.8]$ \\
def compiles   & \pos{$-8.3$} & $[-15.2, -1.4]$ & $-4.6$ & $[-11.5, +3.1]$ & $+4.1$ & $[-3.4, +11.0]$ \\
\bottomrule
\end{tabular}
\end{table}

\paragraph{The gap adapter as a writer.}
Table~\ref{tab:gap} gives the gap adapter's results in the format of Tables~\ref{tab:main} and~\ref{tab:paired}. The one clean positive effect of a hint in the whole study belongs to it: with the oracle hint it compiles $9.0$ more definitions per hundred than without ($[+2.8, +15.2]$). The model trained to consume difference labels is the one that benefits from being handed them, which is a statement about format, not about mathematics.

\paragraph{Is the gain spread across projects?}
Table~\ref{tab:repos} splits the no-hint comparison between the base model and the writer adapter by benchmark project. The adapter has more statements in gold agreement than the base model in 15 projects, the same number in 8, and fewer in 4; for proofs the counts are 12, 10 and 5. No single project drives the result: the three largest contributions (repo-0674, $+4$; repo-0292 and repo-0179, $+3$ each) add up to less than half of the $+22$ statements in gold agreement gained in total.

\begin{table}[h]
\centering
\caption{Per-project counts, no hint: number of tasks ($n$, def/thm/proof), statements in gold agreement, and completed proofs for the base model (b) and the writer adapter (a).}
\label{tab:repos}
\small
\begin{tabular}{l c cc cc @{\qquad} l c cc cc}
\toprule
 & & \multicolumn{2}{c}{Gold agreement} & \multicolumn{2}{c}{Proof} & & & \multicolumn{2}{c}{Gold agreement} & \multicolumn{2}{c}{Proof} \\
Project & $n$ & b & a & b & a & Project & $n$ & b & a & b & a \\
\midrule
0145 & 0/2/2   & 0 & 0 & 1 & 0 & 0564 & 9/5/5   & 1 & 3 & 2 & 1 \\
0179 & 4/19/19 & 4 & 7 & 2 & 6 & 0612 & 3/7/7   & 1 & 2 & 1 & 3 \\
0262 & 6/10/10 & 2 & 3 & 1 & 4 & 0618 & 0/5/5   & 2 & 3 & 1 & 1 \\
0292 & 1/11/11 & 3 & 7 & 0 & 3 & 0621 & 1/12/12 & 3 & 3 & 1 & 0 \\
0421 & 12/14/14 & 3 & 5 & 2 & 5 & 0627 & 7/9/9  & 4 & 5 & 2 & 2 \\
0423 & 2/2/2   & 1 & 1 & 1 & 1 & 0637 & 6/24/24 & 8 & 10 & 0 & 3 \\
0424 & 4/3/3   & 1 & 2 & 1 & 0 & 0642 & 3/7/7   & 4 & 4 & 3 & 3 \\
0463 & 4/2/2   & 0 & 1 & 0 & 0 & 0669 & 10/11/11 & 1 & 1 & 0 & 1 \\
0465 & 9/0/0   & 0 & 0 & 0 & 0 & 0670 & 4/4/4   & 3 & 3 & 0 & 1 \\
0469 & 19/10/10 & 4 & 3 & 2 & 1 & 0674 & 6/24/24 & 7 & 11 & 2 & 4 \\
0470 & 1/7/7   & 1 & 2 & 0 & 0 & 0767 & 5/17/17 & 6 & 5 & 0 & 1 \\
0483 & 5/10/10 & 5 & 3 & 3 & 3 & & & & & & \\
0487 & 7/5/5   & 2 & 2 & 1 & 1 & & & & & & \\
0509 & 5/9/9   & 4 & 3 & 0 & 2 & & & & & & \\
0511 & 9/7/7   & 0 & 1 & 0 & 0 & & & & & & \\
0563 & 3/14/14 & 6 & 8 & 3 & 5 & & & & & & \\
\bottomrule
\end{tabular}
\end{table}

\section{Related work}
\label{app:related}

Autoformalization with large language models was first demonstrated on isolated competition problems~\citep{wu2022autoformalization}, and ProofNet~\citep{azerbayev2023proofnet} extended it to undergraduate textbook theorems, still one statement at a time and with no surrounding project. The proving benchmarks miniF2F~\citep{zheng2022minif2f} and PutnamBench~\citep{tsoukalas2024putnambench} hand the model a formal statement and ask for a proof; this is our proof task, but on self-contained competition problems rather than on declarations inside a research project. Large (natural language, Lean) corpora exist: Lean Workbook~\citep{ying2024leanworkbook} formalizes competition problems at scale, and Herald~\citep{gao2024herald} back-translates Mathlib statements into natural language. Both pair a statement with its translation; neither records how a formalization departs from the paper it follows, which is what our records do. LeanDojo~\citep{yang2023leandojo} extracts theorems from Mathlib and evaluates proving inside it with retrieval, which is the closest setting to ours; the differences are that Mathlib is a single mature library, and that the task there is proving, not writing statements and definitions that must fit a paper. Provers such as DeepSeek-Prover-V2~\citep{ren2025deepseekproverv2}, Kimina-Prover~\citep{wang2025kimina} and Goedel-Prover-V2~\citep{lin_goedel-prover-v2_2025} are trained with compiler feedback on large sets of competition-style problems. We include Goedel-Prover as a baseline; its high compile rate but low Gold agreement rate in Table~\ref{tab:main} illustrates the difference between solving isolated problems and writing inside a project.

\section{What the two adapters were trained on}
\label{app:examples}

Each training example is a pair: an \emph{input} that the model is shown, and an \emph{output} that it is trained to write. The loss is computed only on the output tokens; the input tokens are masked. Below is one real example from each training file, copied verbatim except that very long lines have been broken and the file prefix in the writer examples has been shortened where marked with \texttt{...}.

\subsection{Writer adapter}

The writer adapter is trained on the Lean itself. Its input is the same prompt the benchmark uses (without any hint): the beginning of the actual project file, the informal source, and an instruction. Its output is the declaration the project actually wrote. Below are a statement example and a proof example. The system message is the same for both and is quoted in Appendix~\ref{app:details}.

\paragraph{Statement example: input (user message).}
{\footnotesize\begin{verbatim}
File `CmpProject/Chen/Theorem21.lean` so far:
```lean
import CmpProject.Chen.Defs
import CmpProject.Chen.Telescope
import CmpProject.Chen.Blocks
import CmpProject.Chen.Seams

/-!
# Theorem 2.1 of Chen's paper
... (module docstring) ...
-/

open PowerSeries

namespace Nat.Partition

/-- The `A`-side of `hypothesis_5prime` always has constant term `1`. -/
private lemma hypothesis_5prime_coeff_zero_Aside (A : Set ℕ)
    [DecidablePred (· ∈ A)] (N : ℕ) :
    (1 + ∑ a ∈ (Finset.Icc 1 N).filter (· ∈ A), (X : ℕ⟦X⟧) ^ a).coeff 0 = 1 := by
  ... (proof) ...
```

Informal source:
Source: Introduction, definitions of P(A;n) and Q(B;n) (Partition theorems and
the Chinese remainder theorem)
Statement (paraphrase): The paper defines P(A;n) by allowed part multiplicities
and Q(B;n) by allowed part values, with both counts set to one at n=0.
Informal argument: The paper defines ℕ as the positive integers and assumes B⊆ℕ
and δ_b∈ℕ, so b=0 and δ_b=0 never arise.

Write the Lean 4 statement of `hypothesis_5prime_delta_pos` (theorem) that
formalizes the following informal result. End the statement with `:= by sorry`;
do not prove it. Output only that declaration.
\end{verbatim}}

\paragraph{Statement example: output.}
{\footnotesize\begin{verbatim}
```lean
/-- **`δ b ≥ 1` is forced by `h5`.** If some `b ∈ B` had `δ b = 0`, that factor
would be the empty sum `0`, zeroing the whole `B`-side product at every
coefficient — contradicting the `A`-side's constant term `1`. -/
theorem hypothesis_5prime_delta_pos (A : Set ℕ) [DecidablePred (· ∈ A)]
    (B : Finset ℕ) (δ : ℕ → ℕ) (N : ℕ) (h5 : hypothesis_5prime A B δ N)
    (b : ℕ) (hb : b ∈ B) : 1 ≤ δ b := by sorry
```
\end{verbatim}}

\paragraph{Proof example: input (user message).}
{\footnotesize\begin{verbatim}
File `CollatzShadowing/Shadowing.lean` so far:
```lean
import Mathlib.RingTheory.Ideal.Basic
import Mathlib.Logic.Function.Iterate
import CollatzShadowing.Phantom
import CollatzShadowing.Auxiliary
import CollatzShadowing.Syracuse2Adic

namespace CollatzShadowing

/-- The shadowing congruence `x ≡ y (mod 2^k)` in `ℤ_[2]`, expressed
as `x - y ∈ ⟨2^k⟩`. -/
def PadicCongruentModPow2 (x y : ℤ_[2]) (k : ℕ) : Prop :=
  x - y ∈ (Ideal.span {((2 : ℤ_[2]) ^ k)} : Ideal ℤ_[2])

... (one more proved theorem) ...
```

Informal source:
Source: paper/collatz_spectral_reduction.tex:249-267 (Exact shadowing statement)
(A Spectral Reduction of the Collatz Conjecture via Phantom Orbit Shadowing)
Statement (paraphrase): The exact shadowing lemma assumes a congruence modulo a
power of two in the 2-adic integers and concludes agreement of the first m
Syracuse valuation steps.
Informal argument: The paper-level step writes a congruence n congruent q_w
modulo a power of two and reasons via 2-adic closeness.

Statement to prove:
```lean
/-- `(3 : ℤ_[2]) ^ j` is non-zero (it is a unit). -/
private theorem three_pow_ne_zero_z2 (j : ℕ) : (3 : ℤ_[2]) ^ j ≠ 0 :=
```

Prove the following Lean 4 statement, which formalizes the informal result
below. Output only the proof: what comes after `:=` (start with `by` for a
tactic proof, or give a term). Do not repeat the statement.
\end{verbatim}}

\paragraph{Proof example: output.}
{\footnotesize\begin{verbatim}
```lean
pow_ne_zero _ three_ne_zero_z2
```
\end{verbatim}}

\subsection{Gap adapter}

The gap adapter never sees any Lean. Its input is a short paraphrase of the informal passage plus a one-line goal, and its output is a JSON object: the difference type (one of the 17 labels) together with three English sentences describing what the formalizer did and why. Only the \texttt{type} field is scored.

\paragraph{Input (system message).}
{\footnotesize\begin{verbatim}
You predict how a Lean 4 / Mathlib formalization will differ from the
natural-language mathematical source it formalizes. You are shown only the
source-side situation and the formalization goal; you never see the Lean code.

Respond with a single JSON object and nothing else:
{"predictions": [{"type": ..., "lean_choice": ...,
                  "source_to_lean_difference": ..., "reason_or_pressure": ...}]}

The "type" field must be exactly one of:
- auxiliary_result_or_construction
- case_analysis_expansion
   ... (all 17 labels are listed here) ...
- totalization_or_choice

Most situations produce exactly one prediction. Emit more only when the
situation genuinely forces several independent divergences.
\end{verbatim}}

\paragraph{Input (user message).}
{\footnotesize\begin{verbatim}
Mathematical situation:
The source treats the natural numbers axiomatically through the Peano axioms
rather than choosing a concrete implementation.

Formalization goal:
Represent Peano arithmetic and establish the natural-number axioms in Lean.

Predict the formalization differences.
\end{verbatim}}

\paragraph{Output (what the model is trained to write).}
{\footnotesize\begin{verbatim}
{"predictions":[{"type":"representation_change",
"lean_choice":"Declare Chapter2.Nat as a Lean inductive type with zero and
 successor constructors, then prove the axioms for that concrete type.",
"source_to_lean_difference":"The natural presentation is axiomatic, whereas Lean
 chooses an initial-algebra-style inductive representation whose elimination and
 constructor laws are built into the kernel.",
"reason_or_pressure":"Lean's inductive types give recursion and induction
 principles automatically and avoid carrying an abstract Peano structure through
 every definition."}]}
\end{verbatim}}

\section{Clarifications and sensitivity analyses}
\label{app:review-clarifications}

\paragraph{Scope of the controls.}
The size-matched writer/gap comparison tests training targets, not the value of selecting discrepancy sites. Isolating that value requires equally sized ordinary-declaration data from the same training projects, with matched task mix and training settings. The gap predictor's majority-baseline performance limits its negative result; oracle hints provide stronger evidence, but only for this 17-label taxonomy and prompt format. Neither experiment rules out richer semantic feedback. 
\end{document}